\pdfoutput=1  

\documentclass[letterpaper, 10 pt, conference]{ieeeconf}  
\IEEEoverridecommandlockouts

\usepackage{graphics} 
\usepackage{amsmath} 
\usepackage{algorithm}
\usepackage{algpseudocode}
\usepackage{orcidlink}
\usepackage{tabularx}
\usepackage{makecell}
\usepackage{booktabs}
\usepackage{multirow}
\usepackage{graphicx}
\usepackage{stfloats}
\usepackage{xurl}
\usepackage{hyperref} 
\usepackage{amssymb}
\usepackage{soul}
\usepackage[caption=false,font=footnotesize]{subfig}
\let\labelindent\undefined
\usepackage{enumitem}
\title{\LARGE \bf
Efficient Real-World Online Reinforcement Learning for Robot Manipulation via Centralized Training and Critic Decomposition
}

\author{
Changhao Li\textsuperscript{1,2,*}\,\orcidlink{0009-0006-1754-0283},
Yifang Zhang\textsuperscript{1}\,\orcidlink{0000-0002-7639-4232},
Heng Zhang\textsuperscript{3} \,\orcidlink{0000-0003-4832-9668},
Davide Torielli\textsuperscript{4}\,\orcidlink{0000-0002-9711-3006},
Damiano Gasperini\textsuperscript{1}\,\orcidlink{0009-0003-1384-1829},
\\
Arturo Laurenzi\textsuperscript{1},
Luca Muratore\textsuperscript{1}\,\orcidlink{0000-0002-1265-3370},
Arash Ajoudani\textsuperscript{3}\, \orcidlink{0000-0002-1261-737X},
Nikos Tsagarakis\textsuperscript{1}\,\orcidlink{0000-0002-9877-8237}
\thanks{$^{1}$ HHCM, Istituto Italiano di Tecnologia, Genoa, Italy.}
\thanks{$^{2}$ DIBRIS, University of Genova, Genova, Italy.}
\thanks{$^{3}$ Human-Robot Interfaces and Interaction Lab, Istituto Italiano di Tecnologia, Genova, Italy.}
\thanks{$^{4}$ Cognitive Robotics, TU Delft, 2628CD Delft, The Netherlands}
\thanks{$^{*}$ Corresponding author: {\tt\small changhao.li [at] iit [dot] it}}
}

\begin{document}

\maketitle
\thispagestyle{empty}
\pagestyle{empty}

\begin{abstract}
Real-world online reinforcement learning (RL) provides a promising approach for training robotic manipulation policies directly in the physical world, avoiding the sim-to-real gap and enabling continuous policy refinement through human-in-the-loop interaction. Recent methods have demonstrated sample-efficient learning through human intervention but remain limited to small randomization ranges and encounter challenges with the non-stationarity induced by concurrently training multiple agents.
To address these limitations, we introduce a unified framework that combines centralized training with decentralized execution (CTDE) and a Hybrid Reward Architecture (HRA). This enables multiple actors to share a centralized multi-head critic. Specifically, we decouple continuous Cartesian arm pose control and discrete gripper control into two actor policies optimized under a centralized critic. The critic is decomposed into task and grasp heads, corresponding to the sparse task reward and a potential-based grasping reward, respectively. We accordingly reformulate the critic and actor objectives to exploit the decomposed Q-values while explicitly accounting for the categorical action distribution of the discrete gripper policy.
Experimental results demonstrate that the proposed framework substantially improves both sample efficiency and policy performance. We validate our approach on two robotic arms and a simulated humanoid robot across tennis ball and banana pick-and-place, pot reset, and simulated block relocation tasks under dimension-wise domain randomization, approximately $\mathbf{5\text{-}25\times}$ larger than those considered in prior work.
Compared with a state-of-the-art baseline, our method improves the success rate from 60\% to 80\% on tennis ball pick-and-place, from 60\% to 90\% on banana pick-and-place, and from 25\% to 95\% on simulated block relocation, while also successfully accomplishing a task where the baseline consistently fails. Videos and more details are available at our project website: \url{https://hil-harc.github.io/}.
\end{abstract}

\section{INTRODUCTION}
Robotic manipulation remains a fundamental challenge, particularly for robust and efficient deployment~\cite{ReviewLearningbasedDynamics}. Reinforcement learning (RL) offers a promising alternative to manually designed controllers by learning behaviors through trial-and-error~\cite{shahidLearningContinuousControl2020}. However, policies trained in simulation often degrade after transfer because of discrepancies in dynamics, contact, actuation, and sensory observations~\cite{chenGeneralPurposeSim2RealProtocol2024,bjelonic2025towards,10517611}. These discrepancies are especially consequential in physical interaction tasks, where even minor modeling or perception errors can result in task failure. Consequently, sample-efficient real-world online RL with human-in-the-loop (HIL) intervention has attracted growing interest, including for embodied-AI fine-tuning~\cite{tang2025deep,guo2025improving}, world action model post-training~\cite{li2026hi}. Training directly on the robot avoids reliance on accurate simulation while allowing policies to adapt to real-world dynamics. By combining offline data with HIL corrections, recent approaches learn nontrivial skills from restricted hardware interactions~\cite{lei2025rl,zang2026rlinf,turcatoAutonomousReinforcementLearning2025a}. Nevertheless, expert teleoperation remains time-consuming and human attention is a scarce resource, making sample efficiency and training stability central challenges in real-world online RL. 

\begin{figure}[t] 
    \centering 
    \includegraphics[width=0.85\linewidth]{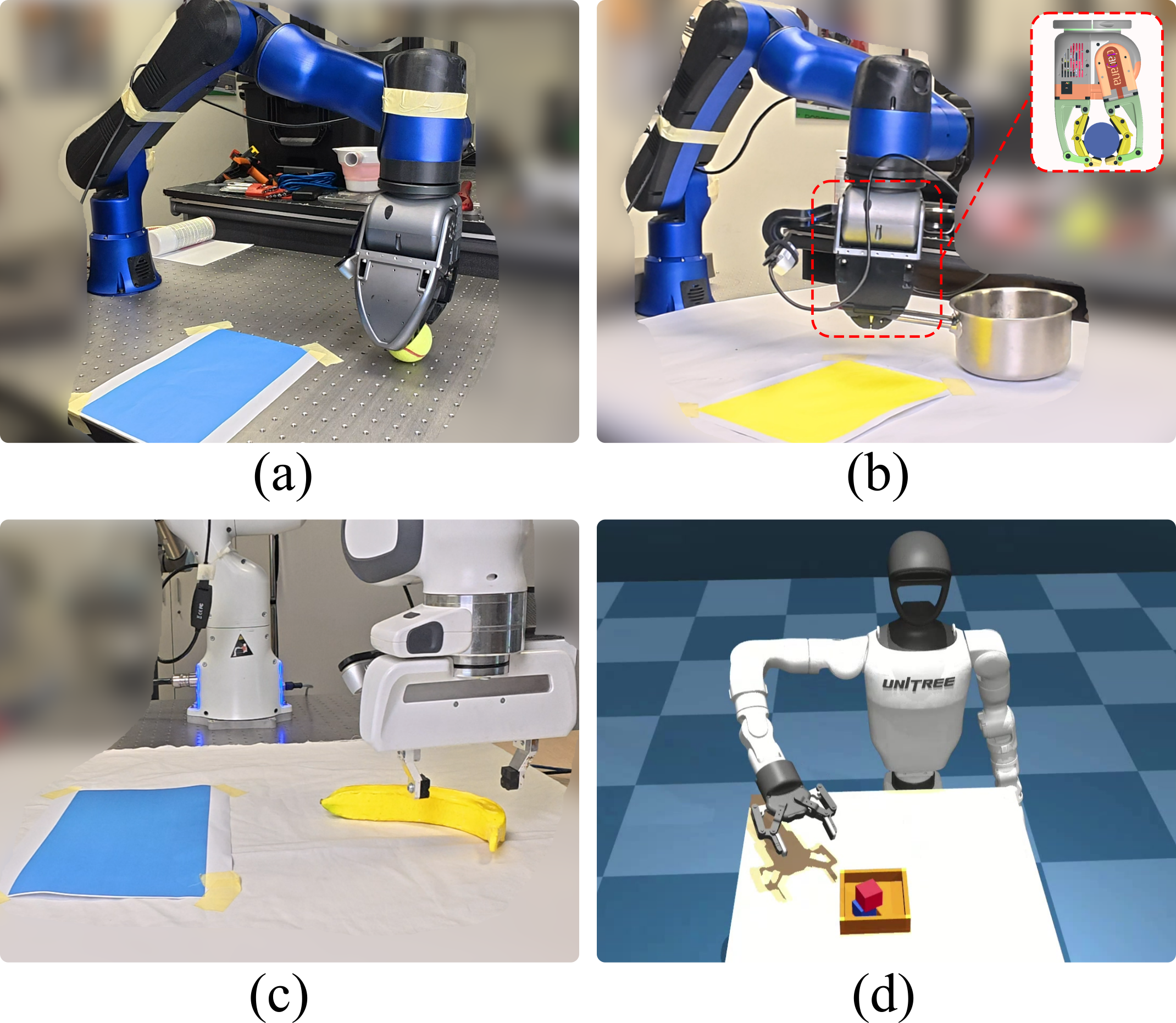} 
    \caption{\textbf{Experimental tasks overview.} (a) A pick-and-place tennis ball task with a custom manipulator; (b) A pick-and-place banana task with a custom manipulator and a compliant gripper; (c) A pot reset task with Franka arm; (d) A block relocation task with Unitree G1 and robotiq gripper} \label{fig:exps} 
    \vspace{-20pt}
 \end{figure}

Despite recent progress, prior methods such as SERL~\cite{luo2024serl}, HIL-SERL~\cite{luo2025precise}, are typically evaluated with limited domain randomization, constrained workspaces, and restricted viewpoints. Similar limitations are also present in online RL fine-tuning vision-language-action (VLA) models~\cite{lu2025vla} and pretrained RL policies~\cite{zhang2026reinflow}. Such bounded exploration and scene diversity hinder policy robustness and generalization, particularly for long-horizon manipulation in more diverse and realistic environments. In this work, we focus on improving policy performance under large-scale domain randomization. Our approach builds upon the HIL-SERL pipeline while introducing several key architectural enhancements to improve robustness across diverse environments, tasks, and robotic platforms, as shown in Fig.~\ref{fig:exps}. 
A second challenge arises from the hybrid nature of manipulation actions, which combine continuous motion with discrete gripper commands. Learning such a hybrid action space with a single policy can introduce non-differentiable decision boundaries, whereas independently optimizing two policies with identical rewards causes each policy to perceive non-stationary transition dynamics as the other policy evolves. This coupling complicates credit assignment and may result in unstable optimization and discontinuous policy updates. Existing discretization or separate Q-learning solutions~\cite{zang2026rlinf,luo2025precise} therefore struggle to represent and optimize both modalities cleanly. Meanwhile, real-world RGB observations are substantially noisier than low-dimensional state representations and simulated views. Their sensitivity to lighting, distractors, and sensor noise can destabilize critic gradients and slow convergence, particularly in sparse-reward settings observed during our deployment.

In this work, we propose a visual manipulation framework designed to improve sample efficiency and policy performance under large-scale domain randomization. We integrate centralized training with decentralized execution (CTDE) for hybrid action learning and a hybrid reward architecture (HRA)~\cite{van2017hybrid} with a centralized multi-head critic. HRA decomposes the original regression target to stabilize learning from high-dimensional noisy observations. We evaluate the framework on both simulation and real-world manipulation tasks, better expose algorithmic capability, providing a comprehensive assessment of robustness and capability. Our contributions are:

\begin{itemize}[leftmargin=*]
\item We introduce a CTDE framework that decouples Cartesian arm pose and gripper control into separate actors sharing a centralized critic. By jointly evaluating both actors' actions, the critic captures arm–gripper interactions, mitigates the non-stationarity in simultaneous multi-agent learning while retaining decentralized execution.

\item We develop an HRA-integrated CTDE formulation with a reward-decomposed multi-head critic. Separate task and grasp value heads transform the original long-horizon value-estimation problem into simpler learning objectives, while the critic and actor losses are reformulated using the decomposed Q-values and the categorical distribution of the discrete gripper policy. This design improves sample efficiency and accelerates convergence under noisy RGB observations.

\item We implement a complete real-world online RL system, including a novel task-adaptive gripper, validated on two robotic arms and across multiple tasks and objects, with additional simulation validation on humanoid manipulation. Results demonstrate improved robustness, training efficiency, and task performance under large domain randomization, supporting practical deployment.
\end{itemize}

\section{RELATED WORK}


\subsection{Real-World Online Reinforcement Learning}
Recent real-world online RL systems such as SERL~\cite{luo2024serl} and HIL-SERL~\cite{luo2025precise} leverage the sample-efficient RLPD framework~\cite{ball2023efficient} for practical on-robot learning under constrained operational settings. RLinf~\cite{zang2026rlinf} extends RLPD with system-level support for training across heterogeneous manipulation and computing platforms. Online RL has also been used to fine-tune vision-language-action (VLA) models~\cite{lu2025vla} and pretrained RL policies~\cite{zhang2026reinflow}. RL-100~\cite{lei2025rl} combines imitation learning, offline RL, and online RL, demonstrating robust real-world deployment with point-cloud observations.

Despite these advances, most systems remain limited to tightly constrained workspaces, often with only a few centimeters of randomization and narrow camera viewpoints, resulting in modest object and scene randomization and limited generalization. RLinf performs RLPD training within small maneuver ranges and restricted variability. Although RL-100 reports comparable trends for RGB and point-cloud observations, our experiments suggest that RGB-based policies are vulnerable to pixel changes, making stable online RL training more challenging. Scaling online RL to larger workspaces further increases the exploration burden: real-world data collection is slow and cannot exploit massively parallel rollouts, while human operator can intervene in only one environment at a time. Moreover, sparse task rewards become increasingly insufficient as workspace size, initial-state diversity, and task horizon increase, further limiting sample efficiency.

Our work scales online RL to wide workspace variations and large randomization. Our framework operates across broad workspace with diverse object configurations and full-range visual views, enabling evaluation under more realistic and demanding conditions. Overcoming constrained randomization and limited viewpoints provides a more robust and generalizable solution for interactive real-world manipulation.

\subsection{Hybrid Action Space with Multi-Agent RL}
Real-world manipulation commonly requires both continuous end-effector motion with discrete gripper commands. Conventional RL policies typically output only one action type, whereas tasks such as pick-and-place require both fine-grained Cartesian control and binary grasping. Prior work, therefore, discretizes parts of the continuous space or relaxes discrete actions into continuous ones~\cite{hausknecht2015deep}. Parameterized action spaces handle hybrid actions without approximation~\cite{xiong2018parametrized,bester2019multi}, while recent actor--critic methods use multi-agent structures with joint critics and separate actors~\cite{lowe2017multi,yu2022surprising}.

However, applying hybrid-action RL to real-world manipulation remains challenging. Existing methods typically either discretize continuous control into a unified action space or optimize heterogeneous policies independently. Discretization reduces control precision and may destabilize optimization near action boundaries, whereas continuous relaxation introduces mapping complexity and can reduce the reliability of binary behaviors such as grasping. Under decentralized training, each agent treats the other agents as part of the environment. Simultaneous policy updates, therefore, change the transition dynamics perceived by each agent, introducing non-stationarity that undermines Markovian process assumptions. These issues hinder deployment in physical environments requiring precise and reliable control.

Our method adopts CTDE~\cite{xiao2022asynchronous,xu2023action,miao2024effective}, using a joint centralized critic with global observations and separate SAC actors for continuous motion and discrete grasping. This design stabilizes multi-agent RL and mitigates non-stationarity while preserving each modality's strengths. The centralized critic also improves training efficiency, which is essential to real-world online RL, providing a robust, practical solution for coordinated continuous and discrete control.

\begin{figure*}[t]
    \centering
    \includegraphics[width=\textwidth]{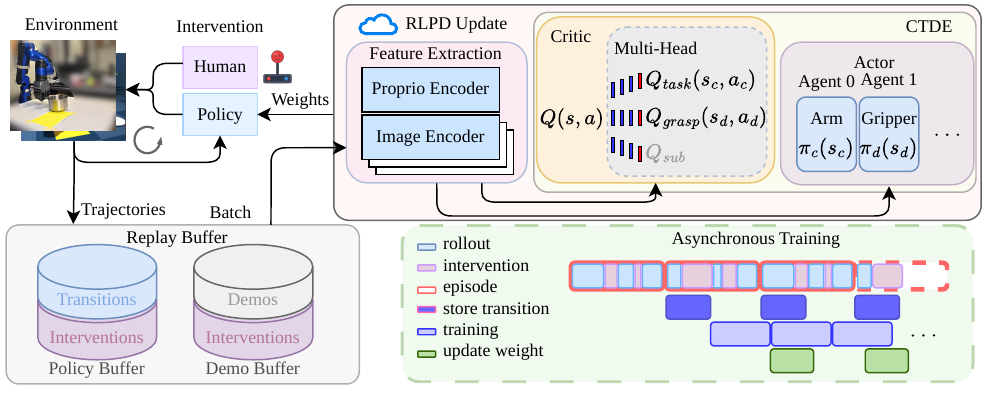}
    \caption{\textbf{Overview of training framework.} Our online RL framework employs a reward-decomposed multi-head critic with multi-agent CTDE training. During training, a human expert provides demonstrations and corrective interventions via teleoperation, while intervened transitions replace the original policy actions and are stored alongside autonomous experience in a shared replay buffer. Training is asynchronous: a cloud-based learner continuously updates the actor--critic, and deployed actors periodically synchronize the latest policy parameters, following the training paradigm of SERL~\cite{luo2024serl} and RLinf~\cite{zang2026rlinf}.}
    \label{fig:traning_framework}
    \vspace{-10pt}
\end{figure*}

\section{METHODOLOGY}
 
The overall framework of our method is illustrated in Fig.~\ref{fig:traning_framework}. It consists of three key components: (1) an RLPD-based real-world online RL pipeline, (2) a multi-agent CTDE training paradigm that mitigates environment non-stationarity, and (3) a reward-decomposed multi-head critic that simplifies value learning through task decomposition. Training follows the RLPD update loop, where the critic, encoder, and actor are sequentially optimized using mini-batches sampled from two replay buffers. The policy buffer stores autonomous interactions and human interventions, while the demonstration buffer contains offline teleoperation data and additional interventions. The demonstration buffer mainly supports initialization and early learning, whereas the policy buffer gradually becomes the primary training source as the policy improves. The framework supports diverse robot platforms, teleoperation devices, cameras, distributed actors, and task-specific rewards, enabling flexible deployment across hardware configurations. Data collection and learning are decoupled: local actors execute policies and collect experience, while actor--critic optimization is performed asynchronously on a remote high-performance workstation with periodic policy synchronization.

\subsection{Sample-Efficient RL with Prior Data}

Consider the RL training a stochastic dynamic system that follows Markov decision process (MDP) $\mathcal{M} = \{\mathcal{S}, \mathcal{A}, p, r, d, \gamma\}$, where $\mathcal{S}$ is the state space, $\mathcal{A}$ is the action space, and $\gamma \in (0,1)$ is the discount factor. The system dynamics are governed by the environment, explicitly by a transition function $p(s'|s, a)$, the same as the reward function r(s, a) and initial state distribution $d(s_0)$. The goal of RL is then to maximize the expected sum of discounted rewards: $E_{\pi}[\sum^{\infty}_{t=0} \gamma^t r(s_t, a_t)]$ to find an optimal policy $\pi$. The policy here is modeled by a Gaussian distribution and parameterized by a neural network.

To enable efficient real-world training, we adopt RLPD~\cite{ball2023efficient}, an off-policy model-free RL framework that leverages prior offline data. Our implementation uses SAC~\cite{haarnoja1812soft} as the underlying learner. Training is performed using a replay buffer $\mathcal{D}$ that combines autonomous policy rollouts, offline expert demonstrations, and human intervention data. All tasks employ sparse task rewards, obtained either from a pre-trained ResNet-based classifier~\cite{he2016deep} or human feedback. Following RLPD, prior demonstrations and online interaction data are sampled from the replay buffer with a predefined ratio, set to 50\% offline data and 50\% online data in our experiments. By incorporating human interventions into the replay buffer, experts can correct undesired behaviors during training, accelerating policy improvement and enhancing final task performance. The effectiveness of human-guided online RL has been demonstrated by HIL-SERL~\cite{luo2025precise}. The critic and actor objectives are defined as follows:

\begin{equation}
 \mathcal{L}_Q(\theta) = \underset{(s,a) \sim \mathcal{D}}{\mathbb{E}}[(Q_{\theta}(s, a) - (r(s, a) + \gamma \underset{s' \sim p}{\mathbb{E}} [V_{\overline{{\theta}}}(s')]))^2]
\end{equation}
\begin{equation}
     \mathcal{L}_\pi(\phi) = -\underset{s \sim \mathcal{D}}{\mathbb{E}}[ \alpha H(\pi_{\phi}(\cdot|s)) + \underset{a\sim \pi_{\phi}}{\mathbb{E}}[Q_{\theta}(s, a)]]
     \label{loss_RLPD}
\end{equation}
where $Q_{\theta}$ denotes the action-value function, and $V_{\overline{\theta}}(s') = \mathbb{E}_{a \sim \pi_{\phi}}[Q_{\theta}(s',a)-\alpha \log \pi_{\phi}(a|s')]$ represents the target state-value function. $H$ denotes the policy entropy. To reduce Q-value overestimation, the Q-value here is an ensemble of multiple Q-functions. In each RLPD update-to-data (UTD) cycle with ratio $n$, the critic is optimized for $n$ gradient steps, while the actor is updated only once, enabling more effective utilization of sampled replay buffer data $\mathcal{D}$. We set the target entropy as $\bar{H}=\frac{1}{2}|\mathcal{A}|$ following convention, and automatically adjust the entropy coefficient $\alpha$ using a temperature network.

\subsection{Multi-agent RL with Hybrid Action Space and CTDE}

CTDE is a multi-agent RL training paradigm that can be broadly categorized into value factorization and centralized critic approaches~\cite{amato2024introduction}. Since our framework is based on the actor--critic algorithm SAC, we adopt the centralized critic formulation introduced by MADDPG~\cite{lowe2017multi}, which also motivates recent actor--critic methods such as MAPPO~\cite{yu2022surprising}. Unlike HIL-SERL, which combines a continuous SAC policy with a discrete DQN gripper controller, we replace DQN with a discrete SAC policy, allowing both actors to share a centralized critic. In the original SAC+DQN design, independently optimized policies perceive other agents as changing parts of the environment, causing non-stationarity and unstable value estimation. The centralized critic addresses this issue by jointly evaluating agent actions during training while preserving decentralized execution. Moreover, a shared critic reduces computational overhead compared with multiple independent critics, enabling more frequent learner updates and improved replay buffer utilization, which is particularly beneficial for real-world online RL. During execution, each actor uses only local observations, while the centralized critic is used exclusively for training.

Our method supports heterogeneous observation spaces across actors. In implementation, the gripper policy uses only wrist-camera images, while the arm policy additionally receives a side-view for object localization. This design can naturally extend to dual-arm humanoid manipulation, where different actors can leverage distinct visual and proprioceptive inputs while sharing a centralized critic. For the single-arm setup, we decompose the policy into two actors: a continuous SAC policy for Cartesian end-effector control and a discrete SAC policy for gripper control. Assuming conditional independence between the continuous action space $\mathcal{A}_c$ and discrete action space $\mathcal{A}_d$, both actions are sampled simultaneously at each control step. The policies are defined as:
\begin{equation}
    \begin{split}
        \pi(a|s) &= \pi^c_{\phi}(a_c|s)\pi^d_{\psi}(a_d|s)\\
         &= \prod_{a_i \in \mathcal{A}_c} \pi^c_{\phi}(a_i | s) \prod_{a_j \in \mathcal{A}_d} \pi^d_{\psi}(a_j | s) 
    \end{split}
\end{equation}
After decoupling, we retain the original objective in Eq. (\ref{loss_RLPD}) to optimize $\pi_{\phi}^c(s)$ and introduce an extra discrete SAC objective to update $\pi_{\psi}^d(s)$. Unlike the continuous policy, the discrete actor does not parameterize a Gaussian distribution. Instead, it directly outputs the logits $z$ of a categorical distribution, which are converted into action probabilities as $\log \pi_{\psi}^{d}(a|s)=\log(\mathrm{softmax}(z))$. Correspondingly, the discrete Q-network outputs Q-values for all possible actions simultaneously rather than estimating the value of a sampled action, improving computational efficiency during optimization. The discrete SAC policy loss is formulated as~\cite{christodoulou2019soft}:
\begin{equation}
     \mathcal{L}_{\pi}(\psi) = \underset{s_t \sim \mathcal{D}}{\mathbb{E}}[ \pi_{\psi}(s_t)^T [\alpha_d \log \pi_{\psi}(s_t) - Q_{\theta}(s_t)]
     \label{loss_discrete_sac}
\end{equation}

\subsection{Decomposed Critic via HRA}

HRA is most beneficial when learning is limited by value estimation rather than representation learning. The pot reset task considered in this work exhibits this characteristic. Using the same visual encoder and HIL-SERL pipeline, we successfully learn a policy in simulation even with Gaussian noise injected into both camera images, indicating that the representation and training pipeline are sufficient under simulated conditions. However, when transferred to the real robot, the same framework fails to maintain a stable success rate within a limited training budget. Without continuous human intervention, performance frequently degrades and rapidly collapses without recovery. We attribute this behavior to the critic. Compared with simulation, the critic exhibits substantially larger gradient norms when trained on real-world observations. Noise in raw RGB images and proprioceptive measurements propagates through the shared encoder into the latent representation, forcing a single monolithic $Q$-function to regress the long-horizon return from noisy features. This significantly increases the difficulty of temporal-difference regression and destabilizes critic optimization. HRA mitigates this through reward decomposition, under the condition that each reward component depends on only a small, largely disjoint subset of the state variables \cite{van2017hybrid}. We separate reward into $r_{task}$ and $r_{grasp}$. $r_{task}$ is the sparse task reward, driven mainly by the front view and the end-effector pose, whereas $r_{grasp}$ is associated with the wrist camera and the end-effector grasp state. The decomposed critics associated with individual rewards produce simpler regression targets while reducing the influence of irrelevant noise. HRA therefore improves critic conditioning under noisy real-world observations by replacing monolithic value estimation with a multi-head critic. Hence, we adopt the following HRA reward decomposition:
\begin{equation}
    \label{HRA reward}
    r(s, a) = \sum_{k=1}^{n}{r_k(s, a)} 
\end{equation}
where $n$ is the number of reward components. In our case, $n=2$ and $r_k \in \{r_{\mathrm{task}},\, r_{\mathrm{grasp}}\}$. The grasp reward provides a dense potential-based reward shaping (PBRS) signal derived from the robot's proprioceptive state. We define a grasp potential $\Phi(s)$ as a clipped min--max normalized function of the grasp torque $\Gamma$ or gripper width $\kappa$, using normalization bounds $x_{\min}$ and $x_{\max}$. The corresponding PBRS reward is:
\begin{equation}
    \Phi(s)=\mathrm{clip}\!\left(\frac{x-x_{\min}}{x_{\max}-x_{\min}},\,0,\,1\right),
    \quad x\in\{\Gamma,\kappa\},
\end{equation}
where the shaping reward is $r_{\mathrm{grasp}}(s,a)=\gamma\Phi(s')-\Phi(s)+P$, and $P$ denotes the gripper-switching penalty.

To exploit the decomposed rewards, we split the centralized critic into one value head for each reward component while sharing a common feature extraction trunk. Let $\theta$ denote the parameters of the complete critic. The original action-value function can then be decomposed as:
\begin{equation}
    \label{HRA Q}
    \begin{split}
        Q_{\theta}(s, a) 
        &= \sum_{k=1}^{n}\mathbb{E}\bigg[\sum_{i=0}^{\infty}\gamma^ir_k(s_{t+i}, a_{t+i}) | s_t = s, a_t = a\bigg]\\
        &= \sum_{k=1}^{n}Q_{\theta}^k(s, a) := Q_{HRA}
    \end{split}
\end{equation}

Instead of learning a single high-complexity value function, each critic head estimates the expected return of one reward component. Since each head regresses a simpler target defined over a compact subset of the task, the resulting TD targets exhibit lower variance, leading to more stable value estimation and improved optimization under high-dimensional real-world visual observations. Each critic head corresponds to one reward channel, while both actors share the centralized multi-head critic and optimize weighted combinations of the two value estimates. The critic loss is defined as:
\begin{equation}
    \begin{split}
        &\mathcal{L}_Q(\theta) = \underset{(s, a) \in \mathcal{D}}{\mathbb{E}}\bigg[\sum_{k}(y^k - Q_{\theta}^k(s, a))^2\bigg]\\
        &y^{\mathrm{task}} =
        r_{\mathrm{task}}+\gamma\left(V_{\mathrm{task}}(s')-\alpha_d\log\pi_\psi(a_d'|s')\right) \\
        &y^{\mathrm{grasp}} =
        r_{\mathrm{grasp}}+\gamma V_{\mathrm{grasp}}(s') \\
        &V_k(s') =\underset{a_d'\sim\pi_\psi}{\mathbb{E}} \left[\min Q^{k}_{\bar\theta_k}(s',a')\right]
    \end{split}
    \label{eq:critic_loss}
\end{equation}
where $V_k(s')$ and $y^k$ denote the state value and TD target of reward head $k\in\{\mathrm{task},\mathrm{grasp}\}$ with parameters $\theta_k$ and target parameters $\bar{\theta}_k$, respectively. The critic evaluates the executed hybrid action $a=(a_c,a_d)$. Accordingly, we replace the Q-value in Eq.~(\ref{loss_RLPD}) and Eq.~(\ref{loss_discrete_sac}) with $\mathbb{E}_{a_d'\sim\pi_\psi}[Q^{\mathrm{task}}+\lambda_gQ^{\mathrm{grasp}}]$ and $Q^{\mathrm{task}}+\lambda_gQ^{\mathrm{grasp}}$, respectively, where $\lambda_g$ controls the contribution of the grasp value in the discrete gripper objective.

Our method, hybrid actors with a reward-decomposed critic (HARC), follows the CTDE paradigm. The continuous arm actor $\pi_\phi$ and discrete gripper actor $\pi_\psi$ act independently during execution but are trained using a shared joint critic that evaluates their combined actions. The critic is decomposed into task and grasp heads, each associated with its own reward component, TD target, and gradient path. The two heads are recombined during actor updates through an adjustable weighting factor, providing a dense auxiliary learning signal to improve sample efficiency. The critics are updated with a high UTD ratio using Polyak-averaged target networks. This design is expected to improve sample efficiency and overall training performance.

\subsection{Compliant Gripper Design}\label{sec:compliant_gripper}

\begin{figure}[htbp]
\centering
\includegraphics[width=\linewidth]{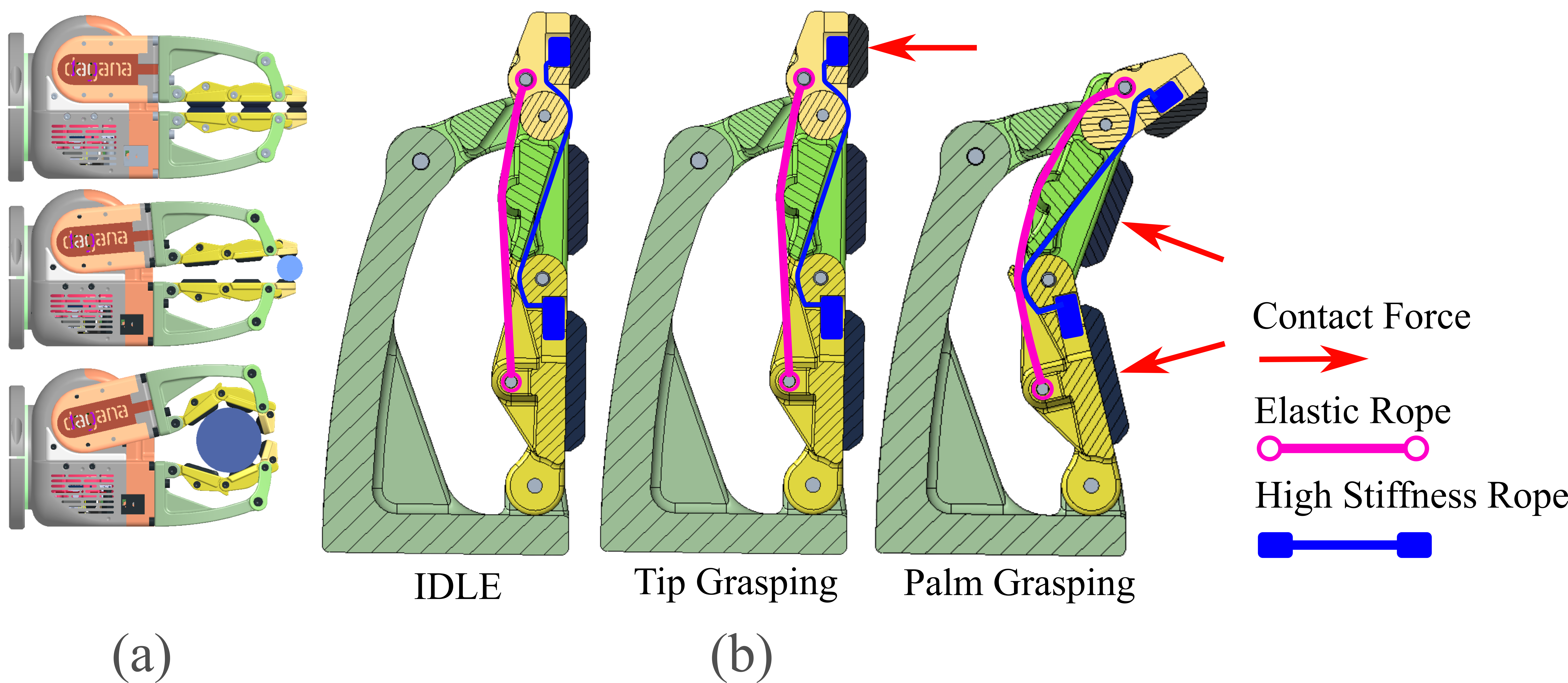}
\caption{Compliant gripper based on the modular 1-DoF revolute joint gripper design~\cite{del2024high}. The left column shows the complete gripper in different configurations, during a fingertip grasp of a small object, and during an enveloping grasp of a larger object, from top to bottom. The enlarged finger diagrams show the extended configuration, the mechanically constrained fingertip-grasp configuration, and the passive adaptation that produces an enveloping grasp.}
\label{fig:compliant_palm}
\vspace{-17.5pt}
\end{figure}

In our experiments, we will deploy the proposed framework to grasp objects with diverse geometric features. Although the standard Franka parallel gripper is effective for objects with simple grasping surfaces, its fixed finger geometry and limited contact area make it difficult to achieve stable grasps on heavier objects and objects with more complex geometries. To overcome these experimental limitations, we developed a compliant two-finger gripper based on the actuation module of a modular 1-DoF revolute-joint gripper~\cite{del2024high}.

As shown in Fig.~\ref{fig:compliant_palm}, the left column presents the complete gripper in three representative configurations: the closed configuration without an object, a fingertip grasp of a small object, and an enveloping grasp of a larger object. The enlarged views illustrate the corresponding passive finger mechanics. Each finger consists of multiple articulated segments coupled by fixed-length, high-stiffness tendons and elastic return elements, indicated by the colored paths in the figure. The tendons coordinate the motion of the finger segments, while the elastic elements restore the finger to its maximally extended idle configuration after contact is removed.

During a fingertip grasp, contact occurs near the distal end of the finger. Mechanical stops limit further relative motion between the segments, allowing the finger to behave approximately as a rigid finger and securely grasp small objects. During an enveloping grasp of a larger object, a larger contact area closer to the palm is required to achieve a secure grasp, and contact is therefore more likely to occur near the finger root. The applied contact forces passively rotate the proximal segments, while the tendon coupling simultaneously bends the distal segment around the object. This adaptation increases the effective contact area and improves grasp stability for larger, heavier, and more geometrically complex objects.

The gripper, therefore, supports both fingertip and enveloping grasps without requiring additional actuators or finger-level control. At the policy level, it retains the same binary open--close interface used in our RL formulation and does not increase the dimensionality of the learned action space.

\section{EXPERIMENTAL SETUP AND RESULTS}

\subsection{Policy Training Configuration}

For each task, RL training is initialized with 20 demonstrations collected via keyboard or joystick. The proposed method is compared with HIL-SERL, a state-of-the-art baseline, under dimension-wise domain randomization approximately $\mathbf{5\text{-}25\times}$ larger than those adopted by HIL-SERL, which uses $2\text{-}8~\mathrm{cm}$ in $x$--$y$ and $1\text{-}10^\circ$ rotational randomization. The randomization configuration is described in Subsection~\ref {sec:experiment_overview}. As reported in SERL, behavioral cloning (BC) with limited demonstrations is insufficient to solve their manipulation tasks without online exploration, and the limitation should be further amplified under large domain randomization due to the increased exploration space. Robot resetting is performed through scripted motions, while environment resetting is manual. The initial end-effector pose is randomized within a $6 \times 6~\mathrm{cm}$ $x$--$y$ region and up to $10^\circ$ $z$-axis rotation. All experiments are conducted on a single NVIDIA RTX 4090 GPU with approximately 2.5 hours of wall-clock training. The environment operates at $10~\mathrm{Hz}$. Policies output 6-DoF Cartesian delta poses with bounded step sizes and a 1-DoF discrete gripper command with three actions: open, close, and stay.

The policy observations including RGB images, end-effector tool center point (TCP) pose, TCP velocity, gripper position, and gripper torque. Since the Franka gripper does not provide torque feedback, torque observations are excluded. Following the baseline, a variable end-effector reference frame is adopted to encourage learning0 relative motions between the end-effector and target objects directly from visual. For fair comparison, both methods use the same RLPD hyperparameters listed on the left side of Table~\ref{tab:rlpd_hysac_params}. Additional hyperparameters on All discrete gripper actors in the proposed method use only wrist-camera inputs. All discrete actors in the proposed method use only wrist camera as the visual input. Despite this restricted observation, the gripper policies successfully learn grasping and releasing behaviors after convergence, enabled by critic decomposition and loss reformulation based on both task and grasp Q-values. RMSNorm~\cite{zhang2019root} is adopted instead of LayerNorm for improved computational efficiency.

\begin{table*}[t]
\centering
\caption{Hyperparameters for RLPD and HARC.}
\vspace{-5pt}
\label{tab:rlpd_hysac_params}
\footnotesize
\renewcommand{\arraystretch}{0.95}
\setlength{\tabcolsep}{3pt}

\begin{tabularx}{\textwidth}{
    @{}lX
    lX
    |lX@{}
}
\toprule
\multicolumn{4}{c|}{\textbf{Shared RLPD Parameters}}
&
\multicolumn{2}{c}{\textbf{HARC Parameters}} \\
\cmidrule(r){1-4}
\cmidrule(l){5-6}

\textbf{Parameter} & \textbf{Value}
& \textbf{Parameter} & \textbf{Value}
& \textbf{Parameter} & \textbf{Value} \\
\midrule

Critic network
& $256 \times 256$
& Actor network
& $256 \times 256$
& critic head network
& $256 \times 256$ \\

Image encoder
& ResNet-10 + MLP
& Proprioceptive encoder
& MLP
& Normalization
& RMSNorm \\

Buffer size
& 100k
& Critic ensemble
& 2
& Discrete actor output
& 3 \\

Critic subsamples
& None
& Discount factor
& 0.97
& Discrete actor network
& $256 \times 256$ \\

Batch size
& 256
& UTD
& 2
& Discrete actor LR
& $3 \times 10^{-4}$ \\

Image encoder output
& 256
& Proprio encoder output
& 64
& Temperature LR
& $3 \times 10^{-4}$ \\

Critic LR
& $3 \times 10^{-4}$
& Actor LR
& $3 \times 10^{-4}$
& Initial temperature
& 0.01 \\

Temperature LR
& $3 \times 10^{-4}$
& Initial temperature
& 0.01
& Target entropy
& 0.1 \\

Target entropy
& $-3$
& Entropy backup
& False
& Discrete actor image
& Wrist camera \\

Target update ratio
& 0.005
& Target update frequency
& 1
& Grasp reward coefficient
& 1.0 \\

Actor update frequency
& 4
& Share encoder
& True
& &
\\

\bottomrule
\end{tabularx}
\vspace{-10pt}
\end{table*}

\subsection {Experimental Setups and Task Design Overview}
\label{sec:experiment_overview}

We evaluate our method against the baseline using two different robotic arms and a simulated humanoid robot: a 6-DOF custom manipulator, the 7-DoF Franka Emika Panda arm, and a Unitree G1, as shown in Fig.~\ref{fig:exps}. The designed tasks are:

\begin{enumerate}[leftmargin=*]
\item \textbf{P\&P a tennis ball:}

The robot grasps a tennis ball and places it on a yellow marker. The ball is randomized within a $50 \times 40~\mathrm{cm}$ workspace. Failed grasps may transition into pick-and-catch as the ball rolls along the gripper closing direction. Observations are provided by a front camera and a wrist camera mounted on a customized gripper, with the wrist view enabling egocentric feedback for reliable control. Unless otherwise specified, all tasks use a maximum episode length of 200 steps.

\item \textbf{P\&P a banana:}

This task uses the same visual observations as the tennis-ball task, but excludes gripper torque. The banana pose is randomized within a $30 \times 30~\mathrm{cm}$ planar region with $360^\circ$ orientation variation, requiring robust accurate grasp-point selection under large shape variations.

\item \textbf{Pot reset:}

The robot grasps a metal pot by its flat oval handle and places it on a marker within 250 steps. The pot is randomized within a $40 \times 40~\mathrm{cm}$ workspace with up to $90^\circ$ orientation variation. Its reflective surface, large pose variation, and unstable handle contacts increase task difficulty. Our compliant palm improves handle grasping reliability beyond the original 1-DoF revolute joint gripper.

\item \textbf{Block relocation:}

The humanoid robot relocates two different-sized blocks into a container. The smaller cube requires fingertip grasping for stable transportation. Block positions are randomized within $20 \times 10~\mathrm{cm}$ and $10 \times 10~\mathrm{cm}$ regions with up to $360^\circ$ orientation variation. The arm policy receives head and wrist camera observations and must handle temporary occlusion of the remaining cube during transportation.

\end{enumerate}

\subsection{Experiment Results}

\begin{figure*}[t]
    \centering
    \includegraphics[width=0.98\textwidth]{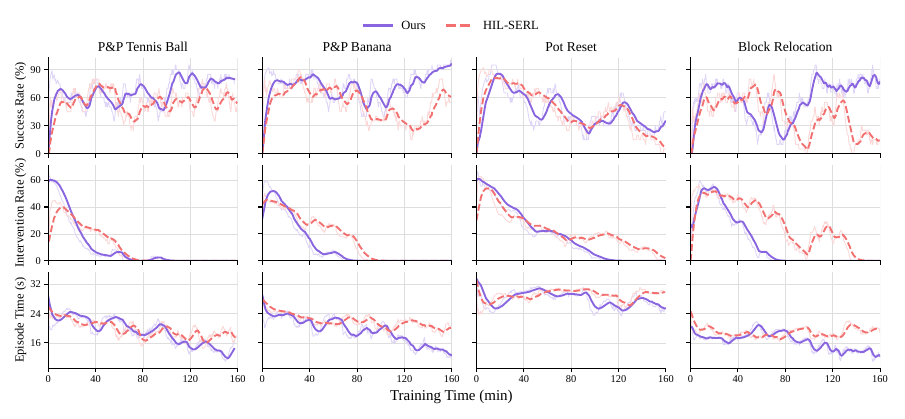}
    \vspace{-5pt}
    \caption{Learning curves for our method and HIL-SERL~\cite{luo2025precise} across all tasks, showing 20-episode running averages of success rate, intervention rate, and episode duration. Success rates increase until convergence or the training-time limit, while intervention rates and episode durations decrease; intervention rates ultimately reach 0\%.
}
    \label{fig:learning_curve_real}
    \vspace{-15pt}
\end{figure*}

\begin{enumerate}[leftmargin=*]
    \item Sample efficiency improvement

    Fig.~\ref{fig:demo_transitions} compares the total expert data required from initialization to training interventions, demonstrating that our method requires less expert supervision than HIL-SERL. Assuming consistent expert demonstrations and converting each initial demonstration transition set to 20 equivalent episodes, the total expert data corresponds to 69, 76, 111, and 115 episodes for our method, compared with 80, 102, 132, and 189 episodes for HIL-SERL. Although the total number of demonstration episodes is not negligible, interventions occur only in short segments when the policy behaves poorly, concentrating expert on improving exploration. This helps to improve sample efficiency and data quality, and enables the policy to learn valuable metrics such as failure recovery. SERL reports that BC performs poorly even with 100 expert demonstrations, while HIL-SERL uses 200 demonstrations to train a diffusion policy. Therefore, our method improves sample efficiency while reducing the need for continuous expert supervision.
    
    \begin{figure}[htbp]
        \setlength{\abovecaptionskip}{0pt}
        \setlength{\belowcaptionskip}{0pt}  
        \centering
        \includegraphics[width=\linewidth]{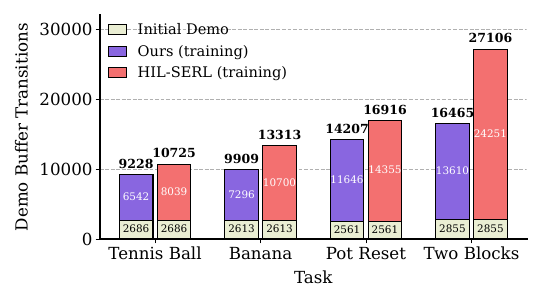}
        \caption{Comparison of demo buffer transitions across tasks. Light green bars indicate the initial 20 episodes of teleoperated demonstrations, while purple and pink bars show additional human-intervention transitions collected during training.}
        \label{fig:demo_transitions}
        \vspace{-10pt}
    \end{figure}
    
    \item Performance under large-scale randomization
    
    Table~\ref{tab:success_rate} reports success rates over 20 evaluation episodes for all tasks, while Fig.~\ref{fig:learning_curve_real} presents the corresponding training curves. After 160 minutes of wall-clock training on the real-world tasks, our method achieves the highest success rate across all benchmarks, increasing the average success rate from 40\% with HIL-SERL to 75\%. On the simulated Unitree block relocation task, our method reaches a 95\% success rate, whereas the baseline achieves only 25\% under the same training budget, highlighting its difficulty in handling multi-object manipulation under large domain randomization. Overall, our method converges faster, requires fewer expert interventions, and completes episodes more efficiently, demonstrating superior task performance, sample efficiency, and execution efficiency under large-scale domain randomization.

\end{enumerate}

\begin{table}[h]
    \centering
    \caption{Policy evaluation performances.}
    {\setlength{\aboverulesep}{0pt}
    \setlength{\belowrulesep}{0pt}
    \begin{tabular}{l | c c}
    \toprule
    \multirow{2}{*}{\textbf{Task}} & \multicolumn{2}{c}{\textbf{Success Rate (\%)}} \\
    \cmidrule(lr){2-3}
     & \textbf{HIL-SERL~\cite{luo2025precise}} & \textbf{Ours} \\
    \midrule
    P\&P tennis ball & 60  & \textbf{80 (+33\%)} \\
    P\&P banana  & 60  & \textbf{90 (+50\%)} \\
    Pot reset & 0 & \textbf{55} \\
    Block relocation & 25 & \textbf{95 (+280\%)} \\
    \bottomrule
    \end{tabular}}
    \label{tab:success_rate}
    \vspace{-10pt}
\end{table}

\section{CONCLUSIONS}

This work presents a practical framework for real-world online reinforcement learning in sparse-reward robotic manipulation tasks with hybrid action spaces and large-scale domain randomization. The framework combines centralized training with decentralized execution (CTDE) and a Hybrid Reward Architecture (HRA), enabling multiple agents to share a centralized critic while optimizing task-specific actors from heterogeneous observations. By leveraging global information during centralized training, CTDE mitigates the non-stationarity and credit assignment challenges inherent to multi-agent learning while allowing each actor to execute independently. The shared critic preserves the actor--critic paradigm of RLPD without introducing independent Q-learning modules, whereas the multi-head critic simplifies value estimation through reward decomposition and reshapes critic and actor objectives for policy optimization. We instantiate the framework using a continuous SAC arm policy and a discrete SAC gripper policy that share the joint critic while operating on local observations. Experimental results on challenging real-world and simulated manipulation tasks demonstrate that the proposed framework consistently outperforms HIL-SERL in task success rate, sample efficiency, expert intervention requirements, and episode execution time. Future work will extend the proposed framework to collaborative dual-arm manipulation on humanoid platforms.







\bibliographystyle{IEEEtran}
\bibliography{bib/ref}

\end{document}